\documentclass[11pt, a4paper, logo, twocolumn]{berkeley}

\usepackage[all]{hypcap}
\usepackage{xspace}
\usepackage[capitalize,noabbrev]{cleveref}
\usepackage{subcaption}
\usepackage{wrapfig}
\usepackage{lipsum}
\usepackage{tabularx}
\usepackage{multirow}
\usepackage{array}
\newcolumntype{Y}{>{\centering\arraybackslash}X}
\usepackage{makecell}
\usepackage[export]{adjustbox}
\makeatletter
\def\mathcolor#1#{\@mathcolor{#1}}
\def\@mathcolor#1#2#3{%
  \protect\leavevmode
  \begingroup
    \color#1{#2}#3%
  \endgroup
}
\makeatother
\usepackage[toc,page]{appendix}

\let\oldtexttt\texttt
\renewcommand{\texttt}[1]{\oldtexttt{\small#1}}

\title{RLDG: Robotic Generalist Policy Distillation via Reinforcement Learning}
\runningtitle{RLDG: Robotic Generalist Policy Distillation via Reinforcement Learning}

\reportnumber{} 

\renewcommand{\today}{December 2024} 

\author[1$\dagger$]{Charles Xu}
\author[1]{Qiyang Li}
\author[1$\dagger$]{Jianlan Luo}
\author[1]{Sergey Levine}

\affil[1]{Department of EECS, UC Berkeley} 
\affil[$\dagger$]{Project Leads}

\correspondingauthor{Jianlan Luo(\href{mailto:jianlanluo@berkeley.edu}{jianlanluo@berkeley.edu}), Charles Xu(\href{mailto:janebair@berkeley.edu}{xuc@berkeley.edu})}

\begin{abstract}
Recent advances in robotic foundation models have enabled the development of generalist policies that
can adapt to diverse tasks. While these models show impressive flexibility, their performance heavily depends on the quality of their training data. 
In this work, we propose Reinforcement Learning Distilled Generalists (RLDG), a method that leverages reinforcement learning
to generate high-quality training data for fine-tuning generalist policies. Through extensive real-world
experiments on precise manipulation tasks like connector insertion and assembly, we demonstrate that
generalist policies trained with RL-generated data consistently outperform those trained with human
demonstrations, achieving up to 40\% higher success rates while generalizing better to new tasks. 
We also provide a detailed analysis that reveals this performance gain stems from both optimized action distributions and improved state coverage.
Our results suggest that combining task-specific RL with generalist policy distillation offers a promising approach for developing more capable
and efficient robotic manipulation systems that maintain the flexibility of foundation models while
achieving the performance of specialized controllers. Videos and code can be found on our project website~\url{https://generalist-distillation.github.io/}.
\end{abstract}

\begin{document}

\maketitle


\begin{figure*}[t!]
    \centering
    \includegraphics[width=1\linewidth]{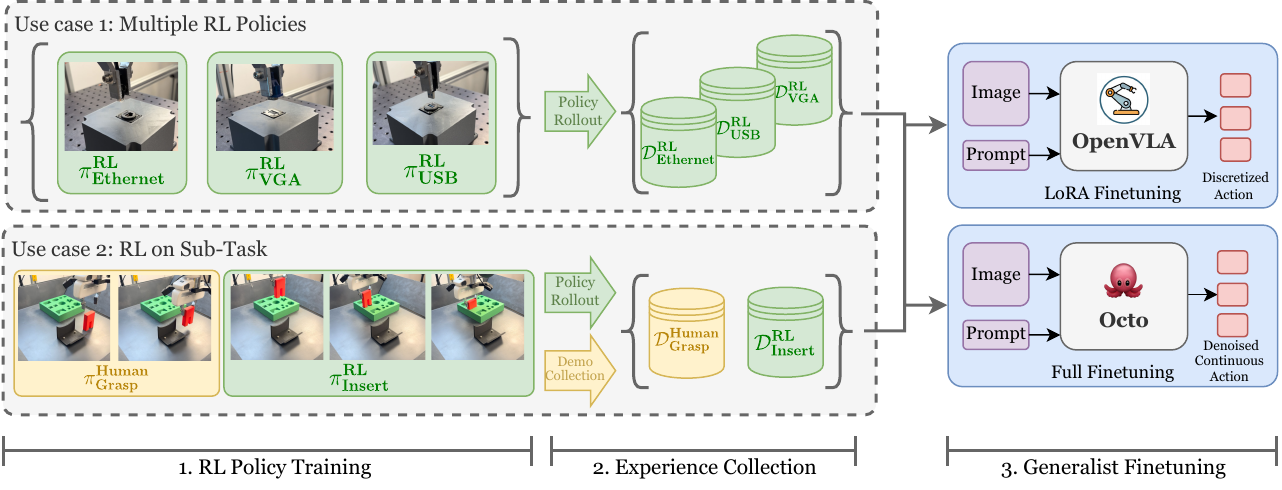}
    \caption{RLDG improves generalist robot policies like OpenVLA and Octo by training with specialist RL policies and using them to generate high-quality fine-tuning datasets. It has the flexibility to distill knowledge from multiple RL policies trained on individual narrowly scoped tasks into a single generalist. It can also be applied to the most critical sub-task of a long-horizon manipulation task, improving the success rate at the ``bottleneck" while leveraging human demonstrations on parts of the task where it suffices.}
    \label{fig:method}
\end{figure*}
\section{Introduction}

Recent advances in robotic foundation models have demonstrated impressive capabilities in understanding and executing diverse manipulation skills~\citep{rtx,rt1,rt2,octo,openvla,pizero, tinyvla,rdt1b, gr2}.
By leveraging Internet-scale pretraining and grounding with robot actions, these models can achieve zero-shot and few-shot generalization across various domains. Deploying these models typically requires fine-tuning them with task-specific data to adapt to the target task or domain.
The quality of this fine-tuning data is therefore critical to the performance of the resulting policies. While human teleoperation is a common and accessible source for such data, human demonstrations often contain inconsistencies in execution quality and style. These variations make it challenging for foundation models to learn robust policies, as they must cope with imperfections and inconsistencies inherent in human demonstrations.
This challenge affects all robotic tasks but becomes particularly pronounced in scenarios requiring precise control and dexterity, such as contact-rich manipulation. These tasks demand fine-grained, reactive control to succeed, making the quality and consistency of demonstration data even more crucial for effective policy learning.


To tackle this challenge, we propose Reinforcement Learning Distilled Generalist (RLDG), a simple yet effective method that leverages reinforcement learning to generate high-quality training data for robotic foundation models.
While directly fine-tuning foundation models with reinforcement learning is possible in principle, it presents significant challenges including optimization instability, computational costs, and potential catastrophic forgetting of pre-trained capabilities, which makes it largely an open problem. Instead, our key insight is that RL agents can autonomously generate high-quality trajectories through reward maximization, making them better suited for fine-tuning generalist policies compared to human demonstrations. The approach is straightforward: we first train vision-based manipulation policies using sample-efficient real-world RL frameworks~\citep{serl,hil-serl} until convergence, then collect data from these policies to fine-tune robotic foundation models.
This procedure is simple and flexible, offering several benefits. First, it provides an automated approach to generate large amounts of high-quality training data without requiring the effort of human teleoperation, which is particularly valuable since autonomous RL training is significantly more cost-effective than collecting human demonstrations. Second, by combining the optimization capabilities of RL with the strong generalization of foundation models, RLDG produces policies that achieve superior performance while generalizing to novel scenarios. Finally, RLDG provides a valuable solution for complex multi-stage tasks by using RL data to address the ``bottleneck" step that hinders the performance of the overall task.

Through extensive experiments across multiple manipulation tasks with well-defined reward functions, we demonstrate that generalist policies like OpenVLA~\citep{openvla} and Octo~\citep{octo} achieve superior performance when finetuned with RL data compared to human demonstrations, specifically for tasks where RL can learn effective controllers. For precise manipulation tasks such as tight-fitting connector insertions presented in Fig.~\ref{fig:tasks}, RLDG achieves 30\% higher success rates on average.
This performance gap widens further when evaluating generalization: policies trained with RLDG demonstrate significantly better transfer to novel scenarios, with on average 50\% higher success rates. Notably, as we will show in Section~\ref{sec:experiment}, achieving comparable performance to RLDG would require 6-10x more human demonstrations.
For complex tasks such as precise insertion, RLDG can achieve perfect success rates (100\%), while policies trained on human demonstrations plateau at 90\% even with significantly more data.


Our key contributions include introducing RLDG, a simple yet effective method that leverages reinforcement learning to generate high-quality training data for fine-tuning pre-trained robotic foundation models, providing an automated alternative to human demonstrations. Through extensive experiments on robotic manipulation tasks, we demonstrate that RLDG achieves 30-50\% higher success rates compared to conventional fine-tuning with human demonstrations while requiring 6-10x less data. Additionally, we show that RLDG can be flexibly combined with human demonstrations in multi-stage tasks, enabling better overall performance by using RL data for critical phases while maintaining the benefits of human demonstrations for other phases.

Our results suggest a promising direction for robotic learning: using reinforcement learning as an automated source of high-quality training data for foundation models. This synergy enables more capable robotic systems that can both execute skills precisely and generalize effectively to new scenarios through natural language instructions while reducing reliance on human demonstration data collection.

\section{Related Work}

Our work bridges reinforcement learning and foundation models for robotics through policy distillation. By combining these approaches, we develop a general technique for training robust robotic policies that leverage both the performance of RL policies and the flexibility of foundation models. Thus, we survey related work across these three key areas and examine their intersections.

\paragraph{Foundation Models for Robotics.}
Recent advances in vision-language foundation models have enabled the development of generalist robotic policies that can understand and execute diverse tasks through natural language instructions. Such models~\citep{rt1,rt2,openvla,octo,pizero,tinyvla,rdt1b,gr2,palme} leverage large-scale pretraining on Internet-scale vision-language data followed by finetuning on robot demonstrations. While these approaches show impressive generalization capabilities across a wide range of tasks, our experiments demonstrate that they often struggle with precise manipulation tasks that require careful alignment and contact-rich interactions (see Section~\ref{sec:experiment}). This challenge comes from fundamental limitations in the demonstration-based learning approach - human demonstrations, while diverse and adaptable, often lack the precision and repeatability needed for contact-rich manipulation tasks.
RLDG addresses this limitation by complementing the semantic understanding of foundation models with the robust behaviors learned through reinforcement learning, enabling precise manipulation while maintaining the flexibility and generalization capabilities of foundation models.


\paragraph{Reinforcement Learning for Robotic Manipulation.} 
Reinforcement learning has been successfully applied to learn complex robotic manipulation skills in the real world through direct interaction with the environment~\citep{serl,hil-serl,levein2016complex,levine16gps,hu23reboot, residualrl,hu2024ibrl, Rajeswaran-RSS-18, visual_residual_rl}. Prior work has demonstrated RL's effectiveness in learning challenging tasks like precision insertion~\citep{Luo-RSS-21, zhao2022insertion, luo2019reinforcement, luo2018deep}, multi-stage assembly~\citep{gupta21mtrf}, and dexterous in-hand manipulation~\citep{hu23reboot}. A key advantage of RL is its ability to discover optimal action distributions through trial-and-error exploration, leading to more robust and efficient policies compared to pure imitation learning~\citep{luo2023rlif}. 
However, RL policies typically struggle to generalize beyond their training distributions, requiring separate policies to be trained for each task variant or environmental condition.

However, while RL policies can achieve exceptional performance on specific tasks, scaling RL to effectively handle the massive datasets used in foundation model training remains challenging. This is primarily due to computational and algorithmic difficulties in scaling up value function learning and policy optimization to process such large quantities of diverse data. As a result, RL approaches often struggle to match the broad generalization capabilities demonstrated by foundation models trained on Internet-scale datasets.
RLDG bridges this gap by combining the strengths of both approaches - using RL to learn optimal behaviors for specific challenging tasks, then distilling these precise capabilities into foundation models while preserving their broad generalization abilities.


\paragraph{Policy Distillation and Knowledge Transfer.} 
The idea of distilling multiple specialized policies into a single more general policy has been explored extensively in the RL and robotics literature~\citep{gps2014}, including methods that use RL to distill into general-purpose neural networks~\citep{rusu2015policy, parisotto2015actor}, methods that employ bidirectional constraints between specialists and generalists~\citep{teh2017distral,ghosh2018}, and methods that focus on continual learning~\citep{rusu2016progressive,schwarz2018progress}.

These approaches have demonstrated that careful distillation can preserve the essential behavioral characteristics of expert policies while potentially adding beneficial properties like improved generalization or reduced computational requirements. While prior work has explored policy distillation in various contexts, our work introduces two key innovations: (1) we show that distilling RL policies into foundation models that leverage large-scale pre-training yields better results than training from scratch, and (2) we demonstrate that for precise manipulation tasks, using RL-generated data for fine-tuning foundation models produces superior performance compared to using human demonstrations, even when high-quality demonstrations are available. Together, these findings establish RLDG as a practical approach for enhancing foundation models with specialized RL capabilities while maintaining their broad generalization abilities.

\section{Reinforcement Learning Distilled Generalist}
\label{sec:method}
Reinforcement Learning Distilled Generalist (RLDG) is a simple yet effective method for enhancing generalist policy performance through the distillation of specialized RL policies. In our system, we train RL policies for individual tasks and then use these policies to generate training data that can be used to fine-tune a single generalist robotic manipulation policy, such as OpenVLA~\citep{openvla} or Octo~\citep{octo}. Although specialized RL policies can achieve high performance on specific tasks,
they often lack zero-shot generalization and robustness to disturbances. Conversely, generalist policies excel at generalization but can struggle to achieve high performance when trained on human demonstrations, for example due to suboptimal data or modality mismatches between human demonstrators and robot policies (e.g., different viewpoints, memory, and task knowledge). RLDG bridges this gap through knowledge distillation, resulting in more performant generalists compared to finetuning on human demonstrations, while demonstrating stronger generalization capabilities compared to the original RL policies. 
This distillation approach through data generation with RL is agnostic to both the choice of RL algorithm and generalist policy architecture, making it flexible to any model choice. Furthermore, it offers flexibility to train and collect data with separate RL policies trained on multiple narrowly scoped tasks (such as one policy for each connector in the \texttt{Connector Insertion} task). We can also elect to train RL on the ``bottleneck" segments of a long-horizon task that require the most precision and benefit the most from RL-generated data, while leaving the less critical parts for humans to demonstrate. This simplifies the RL training complexity, improves data diversity for better generalist performance, and avoids training RL on unnecessarily long-horizon tasks.

\subsection{Online RL Training}
We can formulate each robotic task as a Markov Decision Process (MDP), where the state $s_t$ consists of RGB images and proprioceptive information, and actions $a_t$ represent desired end-effector movements. The policy objective $\pi(a_t|s_t)$ is to maximize the expected discounted return:
\begin{equation}
    J(\pi) = \mathbb{E}_{\substack{s_0 \sim \rho_0 \\ a_t \sim \pi(a_t|s_t) \\ s_{t+1} \sim P(s_{t+1}|s_t,a_t)}} \left[ \sum_{t=0}^T \gamma^t R(s_t, a_t) \right]
    \label{eq:rl-objective}
\end{equation}
where $\rho_0$ defines the initial robot configurations, $P$ represents the system's transition dynamics, and $R: \mathcal{S} \times \mathcal{A} \to \mathbb{R}$ is a reward function encoding the task objectives.

While RLDG is algorithm-agnostic, we implement RLDG using HIL-SERL~\citep{hil-serl} motivated by its sample efficiency and high performance for learning precise real-world manipulation skills from pixel input. It incorporates human interventions with RLPD~\citep{rlpd} to efficiently learn visuomotor policies that consistently achieve 100\% success rate by maximizing \eqref{eq:rl-objective}.



\subsection{Experience Collection}

After training RL experts for each of the tasks provided to RLDG, we collect a high-quality fine-tuning dataset by rolling out the converged policies. Since we transfer knowledge from RL into the generalist policy only through this data, we have the flexibility to mix experience from multiple sources. For tasks that involve separate RL policies per manipulation object like \texttt{Connector Insertion}, we rolled out each policy and constructed a balanced fine-tuning dataset consisting of equal number of episodes per object. In cases where RL is only trained on a segment of the task like \texttt{FMB Assembly}, we combine the RL rollouts with human demonstrations for the remainder of the task. 

\subsection{Generalist Policy Finetuning}

Robot generalist models are often pre-trained on diverse large-scale datasets before being fine-tuned to improve performance its performance on a particular task while preserving the generalization capabilities form the diverse pre-training. In RLDG, we use the data collected as described above to fine-tune these generalist models. Specifically, suppose we have a pre-trained policy $\pi_0$, we fine-tune it with task-specific dataset $D_{(s_t, a_t)}$ with the following supervised learning objective: 
\begin{equation}
    \mathcal{L}(\theta) = -\mathbb{E}_{(s_t,a_t) \sim \mathcal{D}} [\log \pi_\theta(a_t|s_t)]
    \label{eq:finetune-objective}
\end{equation}
We showcase the efficacy of our method by fine-tuning two pre-trained robot generalist models using different action parametrization.

\paragraph{OpenVLA.} OpenVLA~\citep{openvla} is a 7B-parameter vision-language-action model built on Llama 2~\citep{llama2}. It takes a single image as observation input along with a language instruction. It predicts 7-dimensional actions which are discretized into 256 bins per dimension autoregressively using the standard cross-entropy loss.
To fine-tune the model on our RL-generated dataset, we use the public model weights pre-trained on 970 thousand Open X-Embodiment dataset~\citep{rtx} and apply Low Rank Adaptation (LoRA)~\citep{lora}, a popular parameter-efficient fine-tuning method.

\paragraph{Octo.} Octo is another open-source generalist robotic policy, designed to adapt to diverse sensory inputs and action spaces efficiently. Different from OpenVLA, Octo predicts continuous actions with a diffusion head, which excels at modeling multimodal distributions, helpful for imitating human demonstrations~\citep{chi2024diffusion}. To predict an action, the transformer backbone takes in the tokenized observation and goal, then outputs a readout embedding $e$, which is used to condition the denoising process trained on the standard DDPM objective~\cite{ddpm}. We take the pre-trained Octo-Base model, remove its secondary image tokenizer, and mask out the image goal to match our input modalities, and directly fine-tune the remainder of the network on our RL-generated dataset.

\section{Experiment and Results}
\label{sec:experiment}


Our experiments aim to evaluate RLDG in terms of its ability to improve over both imitation learning methods for training generalist policies (in terms of performance), and its ability to improve over more specialized RL policies (in terms of generalization). We use a test suite of tasks that require precise and delicate manipulation, and thus are particularly challenging for imitation learning methods. Specifically, we focus on two main questions: (1) Is the RLDG approach for training generalists using data from RL more effective than the conventional approach of training on demonstration data? (2) Is the generalist policy that results from RLDG training more effective at generalizing than the RL policies used to generate the training data?




\begin{figure}[t]
    \centering
    \includegraphics[width=0.8\linewidth]{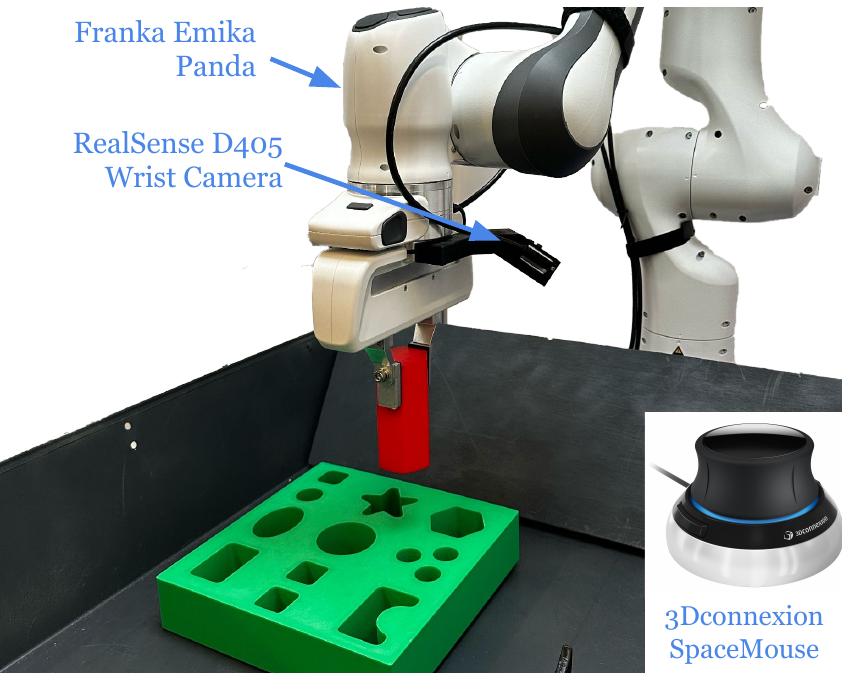}
    \caption{We use a Franka Emika Panda arm with a parallel jaw gripper teleoperated by a 3Dconnexion SpaceMouse device. There is a single RealSense D405 camera mounted on the robot's wrist for image observations.}
    \label{fig:setup}
\vspace{-1em}
\end{figure}

\begin{figure*}
    \centering
    \includegraphics[width=\linewidth]{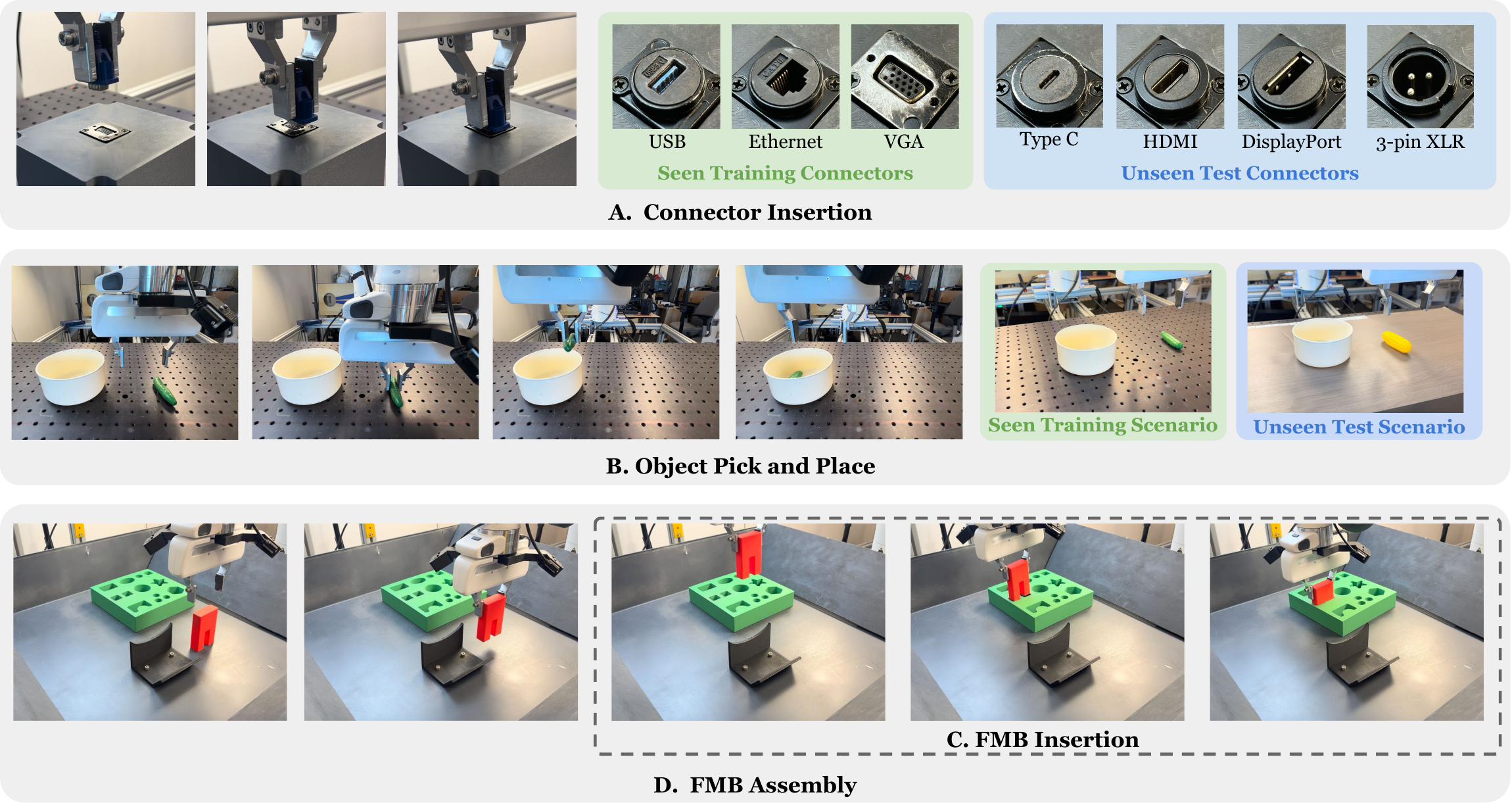}
    \caption{Illustrations of tasks used to evaluate RLDG. (\textbf{A}) Precise \texttt{Connector Insertion} includes three training objects and four unseen test objects for evaluating policy generalization. (\textbf{B}) \texttt{Pick and Place} involves an unseen scenario that tests the policy's visual robustness to different backgrounds and objects. (\textbf{C}) \texttt{FMB Insertion} involves inserting a pre-grasped object in a moving board while (\textbf{D}) \texttt{FMB Assembly} starts with the object on the table and involves an additional grasping phase. }
    \label{fig:tasks}
\vspace{-1em}
\end{figure*}

\subsection{Experimental Setup and Tasks}

\begin{figure*}[h]
    \centering
    \includegraphics[width=1\linewidth]{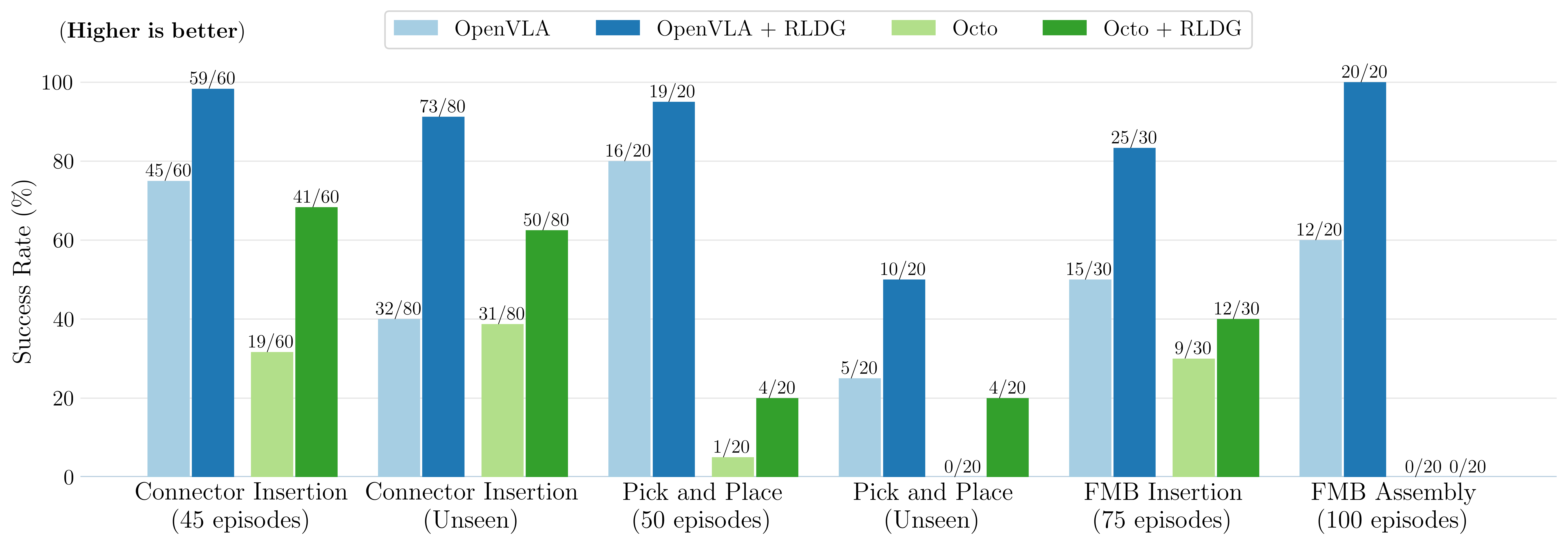}
    \caption{Success rate comparison of OpenVLA and Octo policies fine-tuned with RLDG versus conventional methods using human demonstrations. 
 Both generalists trained with RLDG consistently outperform their counterparts trained with the same number of successful expert human demonstrations in both training and unseen scenarios.}
    \label{fig:rl-vs-human-success-rate}
\vspace{-1em}
\end{figure*}

Our robot setup for all experiments is shown in Fig.~\ref{fig:setup}. The arm tracks end-effector commands with a 1kHz low-level impedance controller. Data collection, RL, and Octo
policies command actions at 10Hz, while OpenVLA runs at 4Hz due to inference speed limitations. The action space for all policies is a 6-dimensional end-effector delta pose in the wrist frame and 1 binary gripper action for tasks that involve grasping.
The RL policy's observation space consists of a single 128 $\times$ 128 wrist RGB image along with end-effector pose, velocity, and wrench measurements. For the generalist policies, we fine-tune only using the wrist camera image as input. 

We evaluate RLDG on four real-world manipulation tasks that present distinct challenges. These include high-precision contact-rich tasks that typically challenge generalist models, pick-and-place tasks where we show RLDG can further improve performance, and multi-stage assembly tasks that leverage RLDG's ability to compose skills. Through these tasks, we also evaluate the method's ability to generalize to unseen configurations. 

\paragraph{Connector Insertion.} This task requires inserting various electronic connectors into their matching ports, which requires sub-millimeter precision
and dexterity to deal with the intricate contact dynamics during alignment. We train separate RL policies and use them to collect data on \texttt{USB}, \texttt{Ethernet}, and \texttt{VGA} connectors before distilling them into a single generalist policy. We also use \texttt{Type-C}, \texttt{HDMI}, \texttt{Display Port}, and \texttt{3-pin XLR} connectors to evaluate the policy's zero-shot generalization performance. 

\paragraph{Pick and Place.} We also test our method on a pick-and-place task, where the robot grasps an object from a randomized location and places it in a bowl. To test generalization, we also evaluate on an unseen scenario by replacing the training object and background as shown in Fig.~\ref{fig:tasks}.
With this experiment, we aim to demonstrate RLDG's effectiveness on tasks that generalist policies are often used for and benchmarked on.

\paragraph{FMB Insertion.} We also use the single object insertion task of FMB~\citep{luo2024fmb}, a common and reproducible benchmark for comparing robotic manipulation methods. This task involves inserting a pre-grasped object into a matching opening with $\pm1.5 mm$ tolerance at randomized positions. We utilize this task primarily for our analysis experiments in Section~\ref{sec:analysis}.

\paragraph{FMB Single Object Assembly.} This task stems from the FMB Single-Object Multi-Stage Assembly, which adds a grasping phase on top of the \texttt{FMB Insertion} task above. We use this multi-stage task to demonstrate RLDG's ability to enhance overall task performance by distilling RL data for the precision-critical insertion phase while using human demonstrations for the grasping and transport phases.

More details on the experiment tasks and training procedure can be found in Appendices ~\ref{sec:appendix-tasks} and ~\ref{sec:appendix-training}.


\subsection{RLDG vs. Conventional Fine-tuning}

In this section, we seek to answer Question 1 by comparing generalist policies fine-tuned using RLDG and standard generalist fine-tuning via imitation learning. For each task, we fine-tune OpenVLA and Octo on RL-generated data as described in Sec.~\ref{sec:method}, and on expert human demonstrations. For a fair comparison, we use the same task setup, training configuration, observation and action space, and the number of successful episodes for both methods. The only difference is the source of the data (RL vs. human). We aim to evaluate whether RL policies are a better source of training data for generalist models than conventional human demonstrations in terms of resultant policy performance.
\paragraph{Success rate.} We present the success rate of each policy and method in Fig.~\ref{fig:rl-vs-human-success-rate}. On each task, both OpenVLA and Octo fine-tuned with RL-generated data consistently achieved higher success rates than their counterparts trained with human demonstrations, in both seen and unseen evaluation scenarios. 
On the precise \texttt{FMB Insertion} and \texttt{Connector Insertion} tasks, where we anticipated the generalist to benefit the most from higher quality training data, OpenVLA with RLDG saw 33\% and 23\% higher success rates, respectively, compared to the baseline. The benefit of RLDG is equally pronounced for Octo, where it improved the success rate by 10\% and 37\%, respectively, although the overall success rate is lower than OpenVLA. We found that RLDG also improved the success rate for \texttt{Pick and Place} from 16/20 to 19/20 for OpenVLA and 1/20 to 4/20 for Octo. Furthermore, the training task performance boost of RLDG also carried over to unseen evaluation scenarios. OpenVLA with RLDG achieved over 2 times higher success rate than OpenVLA with human data for unseen \texttt{Connector Insertion}, while applying RLDG on Octo increased the success rate from 0/20 to 4/20 in the unseen \texttt{Pick and Place} task. When we strategically combine human demonstrations with RL-generated data on the \texttt{FMB Assembly} task, the resulting OpenVLA policy also significantly outperformed the version trained purely on human demonstrations. It achieved 20/20 successes with RL-generated data compared to 12/20 with human demonstrations, suggesting the flexibility and effectiveness of RLDG for improving the ``bottleneck" of a long-horizon task.

\begin{figure}[t]
    \centering
    \includegraphics[width=1\linewidth]{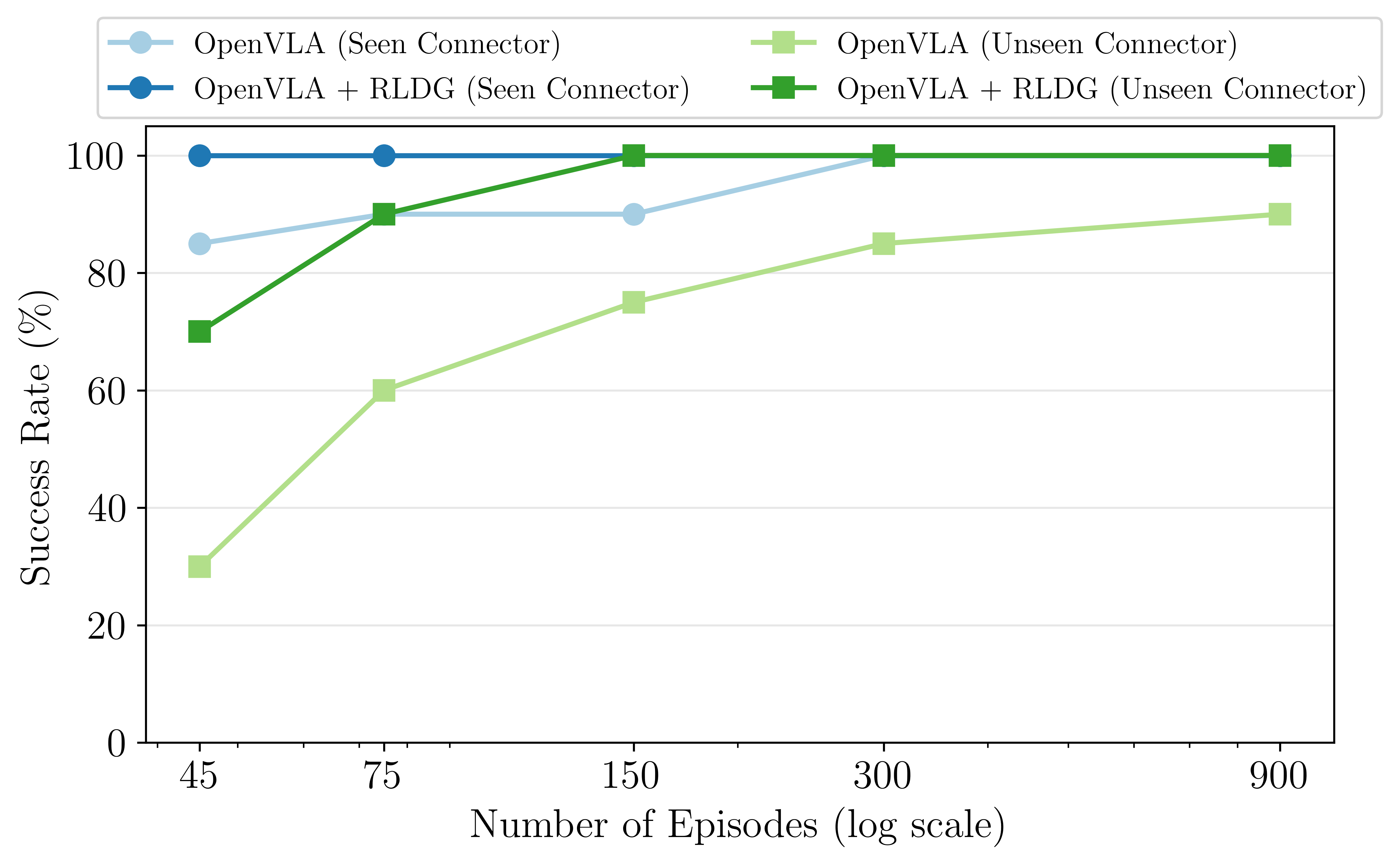}
    \caption{Success rate of OpenVLA policies fine-tuned on different sizes of RL-generated and human-collected datasets. When evaluated on seen (VGA) and unseen (Type C) \texttt{Connector Insertion} tasks, RLDG shows superior sample efficiency, requiring significantly fewer demonstrations to achieve perfect success rate in both scenarios while the performance of conventional method saturates in the unseen case.}
    \label{fig:rl-vs-human-scaling}
\end{figure}
\vspace{-1em}
\begin{figure*}[h]
    \centering
    \includegraphics[width=1\linewidth]{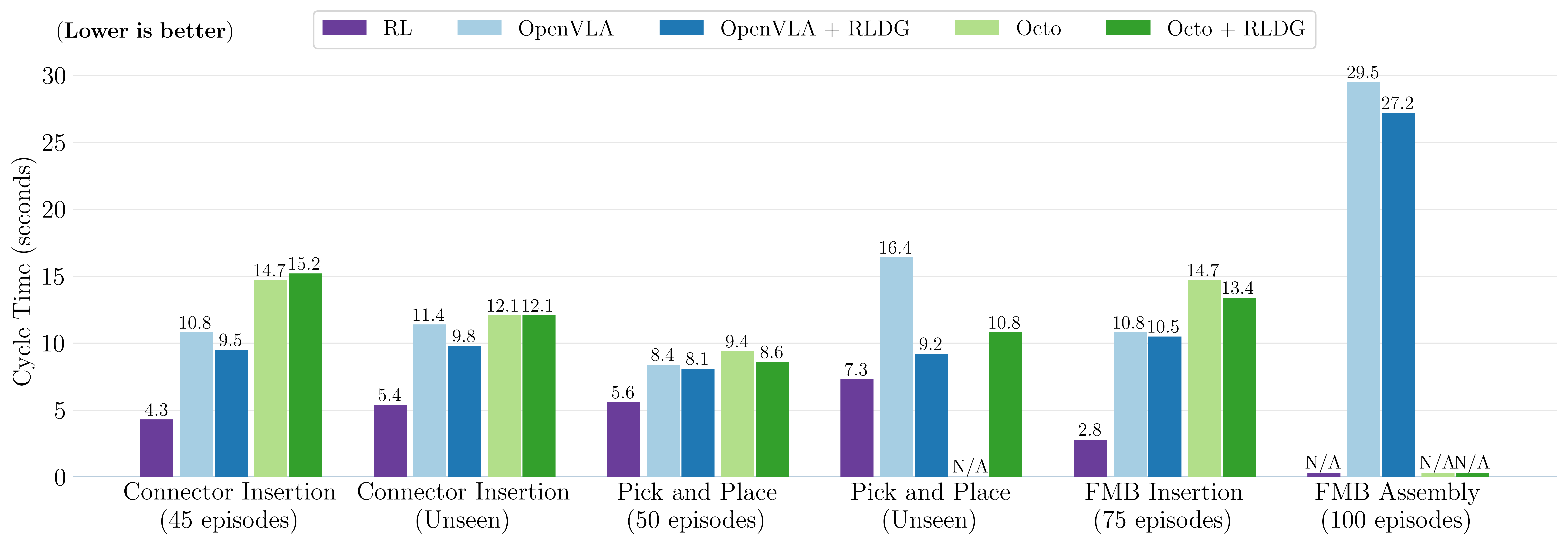}
    \caption{Cycle time comparison between policies trained with RL data versus human demonstrations. N/A for RL in \texttt{FMB Assembly} denotes policy not trained on the whole task, while N/A for fine-tuned policies denotes no successes recorded. The RL-trained policies generally achieve faster execution times across tasks, demonstrating the efficiency benefits of using RL-generated data for policy training.}
    \label{fig:rl-vs-human-cycle-time}
\vspace{-1em}
\end{figure*}
\paragraph{Scaling analysis.}
To further investigate the effectiveness of RLDG, we conduct a scaling experiment studying the success rate of OpenVLA policies on a seen VGA connector and an unseen Type-C connector when fine-tuned on different numbers of RL-generated and human-collected episodes. We present the results in Fig.~\ref{fig:rl-vs-human-scaling}. For the VGA connector, OpenVLA with RLDG achieved a 100\% success rate with just 45 RL episodes, compared to 300 required from human demonstrations to achieve the same success rate. Furthermore, the policy trained with 150 RL rollouts on VGA, USB-A, and Ethernet connectors achieved a 100\% success rate on the unseen Type-C connector insertion, while OpenVLA trained on human demonstrations plateaued at a 90\% success rate even with 900 demonstrations. These results strongly suggest that fine-tuning generalist policies using RLDG is more sample-efficient and leads to higher performance than human demonstrations for both in-distribution and unseen tasks.

\paragraph{Cycle time.} As shown in Fig.~\ref{fig:rl-vs-human-cycle-time}, the generalist policies trained with RL data consistently have faster cycle times compared to those trained with human demonstrations across tasks, although the gap is not as significant. For OpenVLA, RLDG decreased the cycle time between 0.3 to 2.3 seconds per task, while Octo saw little improvement on average. This improvement can be attributed to the inherent speed optimization in RL training through temporal discounting, which is then distilled into the generalist policy by collecting trajectories that solve the task faster than the human expert. However, all generalist policies were still much slower at solving the tasks than the original RL policy. For OpenVLA, we primarily attribute this deficiency to the control frequency gap between the RL policy's 10Hz and OpenVLA's 4Hz,  changing the system dynamics and lowering the maximum velocity of the arm. We believe the speed of OpenVLA can be significantly improved if inference can be sped up to match the RL policy frequency. For Octo, the fine-tuned policies were unable to fit the fine-tuning dataset perfectly, leading to lower success rate and longer cycle time overall. 

\subsection{Generalization of RLDG vs. Original RL Policies}

To address Question 2, we compare the generalization performance of generalists trained using RLDG against that of the original RL policies used to generate the data. As shown in Fig.~\ref{fig:rl-vs-human-success-rate}, the RL policy success rate quickly degraded from 20/20 for the training scenario to 1/20 for the unseen scenario of the \texttt{Pick and Place} task. In contrast, OpenVLA and Octo with RLDG achieved 10/20 and 4/20 success rates respectively on the same task. Additionally, the multi-task capabilities of OpenVLA and Octo allowed fine-tuning on multiple connector data in the \texttt{Connector Insertion} task, achieving 73/80 and 50/80, respectively, when evaluated across 4 unseen connectors, whereas the best of the three RL policies trained on single connectors recorded only 49/80 successes. 


Our experimental results on challenging dexterous manipulation tasks demonstrated several key advantages of RLDG. Generalist robot policies trained on RL-generated data using RLDG consistently achieve higher performance across all tasks compared to conventional fine-tuning methods using human demonstrations. Compared to directly using the RL policies that generated the data, RLDG also demonstrated much greater generalization capabilities and robustness to unseen test scenarios. 

\begin{figure}
    \centering
    \includegraphics[width=\linewidth]{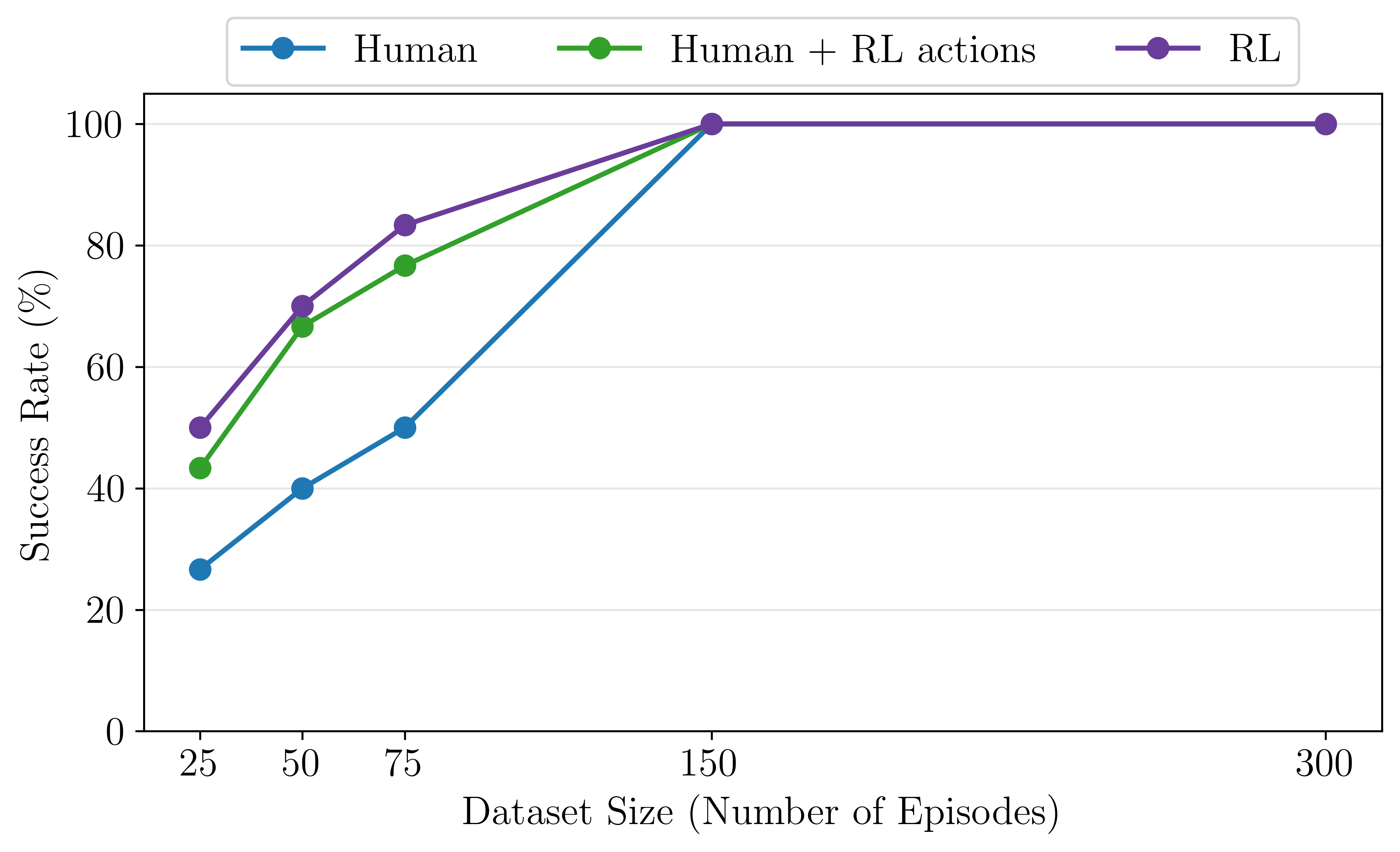}
    \caption{\footnotesize \textbf{Fine-tuning success rate on the \emph{FMB insertion} task with different fine-tuning data sources and varied dataset sizes (from 25 trajectories to 300 trajectories).} \emph{Human}: demo trajectories collected by human teleoperators. \emph{Human + RL actions}: the same human demo trajectories but with all the actions relabeled by a trained RL agent. \emph{RL}: rollouts collected by the RL agent. RL data consistently provide better fine-tuning performance than human data. \emph{Human + RL actions} closes the gap mostly, suggesting that most of the benefits of RL data come from it having better action quality.}
    \label{fig:rl-relabel}
\end{figure}


\begin{figure}[t]
    \centering
    \includegraphics[width=\linewidth]{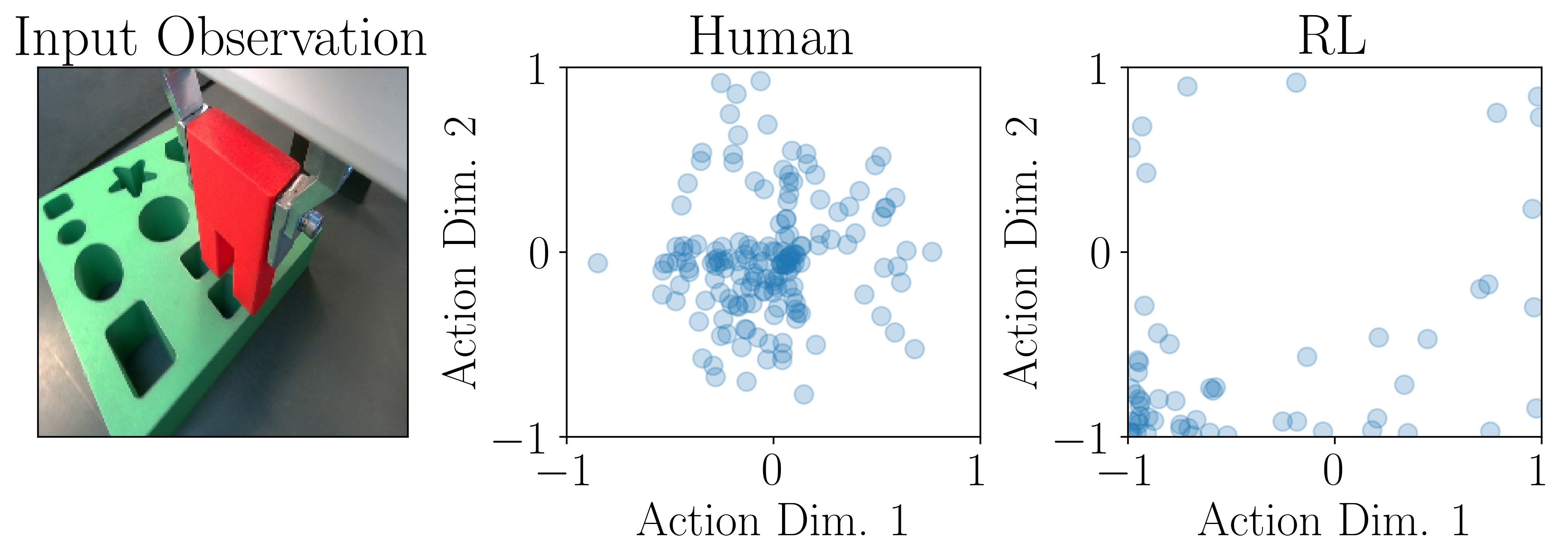}
    \caption{\footnotesize \textbf{Action distribution visualization for RL data and human demo data for the \emph{FMB insertion} task.} We visualize the first two dimensions of the dataset actions after filtering all the transitions in the dataset where the end-effector positions are close to the position shown in the image on the left ($x$/$y$ coordinates are both within $4$mm and $z$ coordinate is within $10$mm). The robot arm needs to move in the -$x$ direction and in the -$y$ direction to reach the insertion point. The first two dimensions of the action space corresponds to the control of the $x$ and $y$ position of the end-effector position correspondingly. Human actions are clustered around the center of the action space whereas the RL actions are more optimized, and mostly found near the correct corner (bottom-left) of the action space.}
    \label{fig:rl-optimal}
\end{figure}


\section{Analysis: why is RL data better than human data?}
\label{sec:analysis}
We have shown that fine-tuning generalist policies with RL data yields superior performance compared to training on human data.
However, it is unclear where these benefits are coming from. 
In this section, we analyze the source of the benefits in two parts. The first part focuses on studying the benefits of RL actions and the state distribution in RL data in isolation. The second part focuses on dissecting the failure modes of the fine-tuned policies on each individual task. 
\subsection{Is RL data better because of better action or state distribution?}
To answer this question, we use the \emph{FMB insertion} task and create a ``mixed'' dataset where we take the human data and relabel the actions using action samples from the RL policy. Comparing the fine-tuning performance of the ``mixed'' dataset with the purely RL data and the purely human demo data would allow us to see the benefits of the actions and the state distribution in isolation.
As shown in Figure~\ref{fig:rl-relabel}, mixing human states and RL actions yields a better fine-tuning success rate than using fully human data (more than 50\% improvements when fine-tuning on 25/50/75 trajectories), while still being worse than using fully RL data. 
This suggests that while RL action and state distribution \emph{both} contribute to the fine-tuning performance improvements, action quality is the factor that contributes to the performance improvement the most. 
Figure~\ref{fig:rl-optimal} shows a comparison of human and RL actions. We can see that the RL action distribution assigns more density on the correct direction
(bottom-left) that moves the end-effector towards the insertion point whereas human action distribution focuses mostly around the middle with a slight bias towards the correct direction. This suggests that RL actions are more optimal than human actions, resulting in the better sample efficiency for fine-tuning we observe in Figure~\ref{fig:rl-relabel}. 
\subsection{Qualitative Analysis: Failure Modes}
\label{sec:failure-modes}
To further understand why RL-generated data leads to better performance, we also analyzed failure modes across tasks.
We observed that policies trained with RL-generated data consistently helped overcome alignment issues in precise, contact-rich tasks and reduced premature gripper closure during grasping. Videos of each policy on each task can be found on our project website (\url{https://generalist-distillation.github.io/}).


\paragraph{Connector and FMB Insertion.} In both tasks, RL data eliminated a ``stuck" state where the object contacts the board but fails to align properly. Human demonstration policies often maintained contact pressure without necessary exploratory movements. Furthermore, RL data also improved approach trajectories, preventing early descents that caused connectors to catch on socket lips. 

\paragraph{Pick and Place.} RL data improved grasp reliability, reducing premature gripper closure seen in human demonstration-trained policies. However, an interesting RL-specific failure mode was observed: objects were sometimes dropped too early, bouncing out of the bowl. This likely resulted from RL's speed optimization, where objects were released immediately after clearing the bowl's edge, but the distilled policy lacked precise timing.

\paragraph{FMB Assembly.} While both OpenVLA policies performed similarly in grasping and transport phases as they are trained on the same human data, the performance gap emerged during insertion, with RL data better addressing alignment issues much like in the insertion tasks. Octo's failure was due to consistent grasping errors where the fingers are in front of the object, likely due to the lack of good depth perception.
\section{Discussion and Limitations}
In this work, we presented RLDG, a simple method that fine-tunes generalist policies on high-quality data generated by RL policies. We demonstrated that generalist policies fine-tuned using RLDG consistently outperform those trained with human expert demonstrations on a suite of real-world challenging precise manipulation tasks. Our method can be applied in real-world robotic manipulation tasks that require a large amount of training data or where policy performance using human demonstrations saturate. Our work also opens up avenues for making autonomous improvements of generalist policies more scalable and tractable. First, our method assumes access to reward functions for fine-tuning tasks which may present difficulties when the task rewards are hard to specify. Possible future directions include autonomously generating fine-tuning tasks with reward functions (e.g., using pre-trained VLMs) such that there is no need for manual task specification. Furthermore, our RL policies optimize not only for task success but the speed in doing so. Such an objective does not necessarily result in policies that are robust in distillation errors. For example, on the \texttt{Pick and Place} task, the policy fine-tuned on RL-generated data always tries to place the object immediately after it has moved close enough to the goal location but sometimes drops the object too early (see Section \ref{sec:failure-modes}). Nevertheless, we demonstrated that specialist RL policies can be an effective generator of training data for robotic foundation models, and we hope to inspire further research in this domain.


\section*{Author Contribution}

\hspace{0.5em} \textbf{Charles Xu} contributed to the research design, prepared the hardware setup, performed the implementation and robot experiments, wrote part of the paper, created the paper figures and the website, co-led the project.

\textbf{Qiyang Li} contributed to the research design, performed the analysis in Section 5, wrote part of the paper

\textbf{Jianlan Luo} conceived the project, designed the research, advised the project in terms of overall direction, practical implementation, experiment design, wrote part of the paper, co-led the project

\textbf{Sergey Levine} conceived the project, designed the research, advised the project in terms of overall direction, experiment design, edited the paper

\section*{Acknowledgments}
This research was supported in part by ONR N00014-22-1-2773, N00014-20-1-2383 and NSF IIS-2150826.

\balance

\bibliography{main}

\newpage

\appendix
\onecolumn
\section{Task Details}
\label{sec:appendix-tasks}

\paragraph{Connector Insertion.} The robot starts with the male connector pre-grasped in a $10 cm \times 10 cm$ plane 5 cm above the female port. We use the same D-type fixture for the female portion which is fixed to the table at a consistent pose. The task is completed if the robot aligns and inserts the connector into the socket. We train the policies on \texttt{USB}, \texttt{Ethernet}, and \texttt{VGA} connectors, while using  \texttt{Type-C}, \texttt{HDMI}, \texttt{Display Port}, and \texttt{3-pin XLR} connectors to evaluate generalization. 

\paragraph{Pick and Place.}In this task, the robot starts $15 cm$ above the table at a fixed pose. The target object is randomly placed within a $18 cm \times 18 cm$ area on the table and the bowl is at a fixed pose $5 cm$ from the edge of the object randomization region. The robot is to pick up the object and put it into the bowl. To test generalization, we evaluate on an out-of-distribution scenario by replacing the green pepper training object with a yellow corn and changing the tabletop to a beige wood grain surface as shown in Fig.~\ref{fig:tasks}.

\paragraph{FMB Insertion.} The robot starts with the insertion object pre-grasped 15 cm above the assembly board, which is randomly placed within a $35 cm \times 35 cm$ area with $\pm15^{\circ}$ rotation. The task is successfully completed if the object is completely inserted into the board. The insertion tolerance in this task is $\pm1.5mm$. We used the same set of board poses during each evaluation experiment to ensure consistency across rus. 

\paragraph{FMB Single Object Assembly.} In this task, the assembly board is randomized the same way as in the \texttt{FMB Insertion}, but the object is randomly placed in a $3 cm \times 7 cm$ grasping area approximately $20 cm$ from the insertion area. The robot starts $15 cm$ above the object at a fixed position. We also use consistent object and board poses during evaluation.

\section{Training Procedures}
\label{sec:appendix-training}

For \texttt{FMB Insertion}, \texttt{Connector Insertion}, and \texttt{Pick and Place}, we first collect positive and negative samples using our SpaceMouse teleoperation device to train a binary success classifier as the reward function for RL. Then, we initialize the RL buffer with 20 demonstrations and train it on the whole task for 1-3 hours until the policy reaches 100\% success rate. Next, we roll out the converged RL policy to collect data for generalist policy fine-tuning, filtering out failed trajectories if any exist. We also collect a set of all successful expert human demonstrations to compare against the RL-generated data. 

For \texttt{FMB Single Object Assembly}, we use the same reward function as in \texttt{FMB Insertion} and we only train RL on the insertion stage by starting the episode with the object pre-grasped and the arm $15 cm$ above the board within a $5 cm \times 5cm$ randomization region. We periodically adjust the grasp pose during training and data collection to build robustness to variations in the grasp. We also collect the same number of human demonstrations for the insertion stage to compare performance. We then separately collect human demonstrations for the grasping phase, starting the robot above the insertion object and ending the episode when the arm has grasped the object and moved within the insertion randomization region above the board. We combine data from the two stages for fine-tuning. 

We fine-tuned the OpenVLA weights pre-trained on OXE dataset using LoRA with a rank of 32 applied to each linear layer of the model, which reduces the computational overhead while not sacrificing much performance. We trained using the default fine-tuning configuration with batch size 2 and 3 gradient accumulation steps on a single Nvidia RTX 4090 GPU, which took between 3 to 5 hours to converge. For Octo, we started from the pre-trained Octo-Base model, using the primary image tokenizer to tokenize the wrist camera image and removed the secondary tokenizer. We also mask out the image goal since we do not use it. We applied full fine-tuning with the default hyperparameters and batch size 64 until convergence, which took 3-5 hours on a single Nvidia RTX 4090 GPU.

\end{document}